\documentclass[journal]{IEEEtran}
\ifCLASSINFOpdf
\else
   \usepackage[dvips]{graphicx}
\fi
\usepackage{url}
\usepackage{graphicx}
\usepackage{xcolor}
\usepackage{mathrsfs}
\usepackage{amsfonts,amssymb}
\usepackage{hyperref}

\begin{document}

\title{AnomalyHop: An SSL-based Image Anomaly Localization Method}

\author{Kaitai Zhang$^\dagger$, \IEEEmembership{Student Member, IEEE},
Bin Wang$^\dagger$, \IEEEmembership{Student Member, IEEE}, 
Wei Wang, \IEEEmembership{Student Member, IEEE}, 
Fahad Sohrab, \IEEEmembership{Student Member, IEEE}, 
Moncef Gabbouj, \IEEEmembership{Fellow, IEEE} and 
C.-C. Jay Kuo, \IEEEmembership{Fellow, IEEE}
\thanks{}
\thanks{Kaitai Zhang, Bin Wang, Wei Wang and C.-C. Jay Kuo are with 
the Ming Heish Department of Electrical and Computer Engineering, 
University of Southern California, Los Angeles, CA 90089 USA (e-mail: kaitaizh@usc.edu).}
\thanks{Fahad Sohrab and Moncef Gabbouj are with Department of Computing 
Sciences, Tampere University, Tampere, Finland.}
\thanks{$^\dagger$ Equally contributed to this paper and listed as co-first author.}
}


\maketitle

\begin{abstract}

An image anomaly localization method based on the successive subspace
learning (SSL) framework, called AnomalyHop, is proposed in this work.
AnomalyHop consists of three modules: 1) feature extraction via
successive subspace learning (SSL), 2) normality feature distributions
modeling via Gaussian models, and 3) anomaly map generation and fusion.
Comparing with state-of-the-art image anomaly localization methods
based on deep neural networks (DNNs), AnomalyHop is mathematically
transparent, easy to train, and fast in its inference speed. Besides, its
area under the ROC curve (ROC-AUC) performance on the MVTec AD dataset
is 95.9\%, which is among the best of several benchmarking methods.
\footnote{Our codes are publicly available at
\href{https://github.com/BinWang28/AnomalyHop}{Github.}}

\end{abstract}

\begin{IEEEkeywords}
Image anomaly localization, successive subspace learning, 
\end{IEEEkeywords}

\IEEEpeerreviewmaketitle

\section{Introduction}\label{sec:introduction}

\IEEEPARstart{I}{mage} anomaly localization is a technique that
identifies the anomalous region of input images at the pixel level.  It
finds real-world applications such as manufacturing process monitoring
\cite{scime2018anomaly}, medical image diagnosis
\cite{schlegl2017unsupervised, schlegl2019f} and video surveillance
analysis \cite{napoletano2018anomaly, saligrama2012video}.  It is often
assumed that only normal (i.e., anomaly-free) images are available in
the training stage since anomalous samples are few to be modeled
effectively rare and/or expensive to collect.  

\begin{figure}[t]
 \centering
\includegraphics[width=0.9\linewidth]{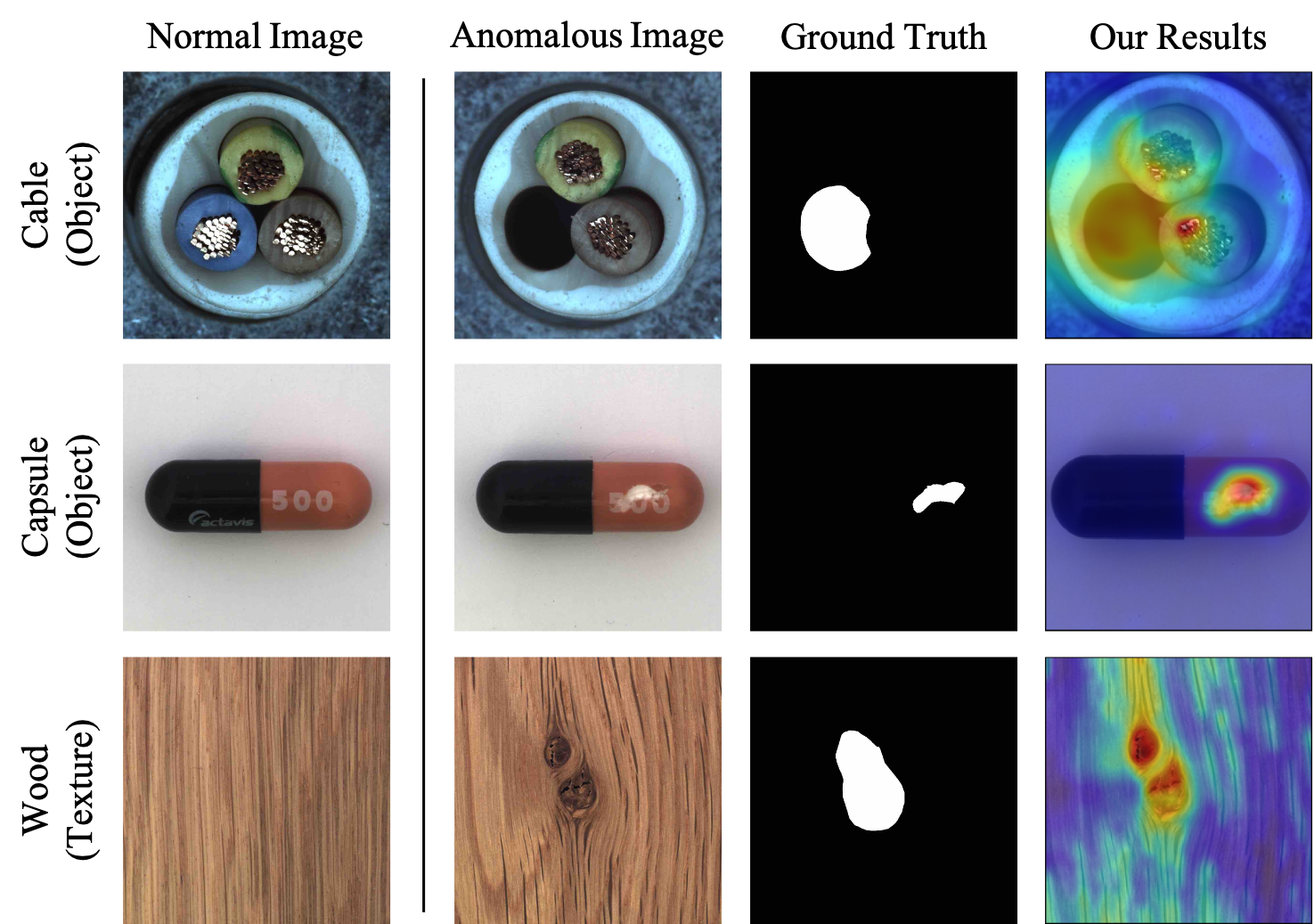}
\caption{Image anomaly localization examples taken from the MVTec AD
dataset (from left to right): normal images, anomalous images, the
ground truth and the predicted anomalous region by AnomalyHop, where the
red region indicates the detected anomalous region.}\label{fig:1_images}
\end{figure}

There is a growing interest in image anomaly localization due to the
availability of a new dataset called the MVTec AD
\cite{bergmann2019mvtec} (see Fig.  \ref{fig:1_images}).
State-of-the-art image anomaly localization methods adopt deep learning.
Many of them employ complicated pretrained neural networks to achieve
high performance yet without a good understanding of the basic problem.
To get marginal performance improvements, fine-tuning and other minor
modifications are made on a try-and-error basis. Related work will be
reviewed in Sec. \ref{sec:review}. 

A new image anomaly localization method, called AnomalyHop, based on the
successive subspace learning (SSL) framework is proposed in this work.
This is the first work that applies SSL to the anomaly localization
problem.  AnomalyHop consists of three modules: 1) feature extraction
via successive subspace learning (SSL), 2) normality feature
distributions modeling via Gaussian models, and 3) anomaly map
generation and fusion.  They will be elaborated in Sec.
\ref{sec:method}.  As compared with deep-learning-based image anomaly
localization methods, AnomalyHop is mathematically transparent, easy to
train, and fast in its inference speed. Besides that, as reported in Sec.
\ref{sec:experiments}, its area under the ROC curve (ROC-AUC)
performance on the MVTec AD dataset is 95.9\%, which is the
state-of-the-art performance.  Finally, concluding remarks and possible
future extensions will be given in Sec.  \ref{sec:conclusion}. 

\begin{figure*}[t]
 \centering
\includegraphics[width=18cm]{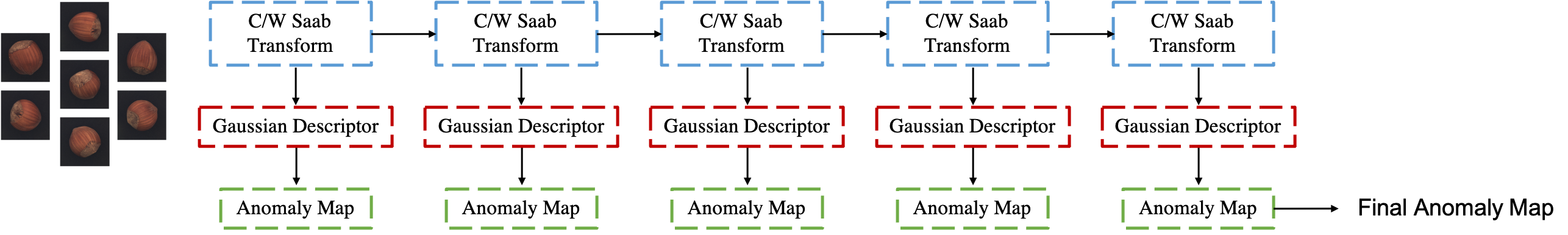}
\caption{The system diagram of the proposed AnomalyHop method.}\label{fig:2_framework}
\end{figure*}

\section{Related Work}\label{sec:review}

If the number of images in an image anomaly training set is limited,
learning normal image features in local regions is challenging.  We
classify image anomaly localization methods into two major categories
based on whether a method relies on external training data (say, the
ImageNet) or not. 

\noindent
{\bf With External Training Data.} Methods in the first category rely on
pretrained deep learning models by leveraging external data. Examples
include PaDiN \cite{defard2020padim}, SPADE \cite{cohen2020sub}, DFR
\cite{yang2020dfr} and CNN-FD \cite{napoletano2018anomaly}. They employ
a pretrained deep neural network (DNN) to extract local image features
and, then, use various models to fit the distribution of features in
normal regions.  Although some offer impressive performance, they do
rely on large pretrained networks such as the ResNet \cite{he2016deep}
and the Wide-ResNet \cite{zagoruyko2016wide}. Since these pretrained
DNNs are not optimized for the image anomaly detection task, the
associated image anomaly localization methods usually have large model
sizes, high computational complexity and memory requirement. 

\noindent
{\bf Without External Training Data.} Methods in the second category
exploit neither pretrained DNNs nor external training data.  They learn
local image features based on normal images in the training set. For
example, Bergmann {\em et al.} developed the MVTec AD dataset in
\cite{bergmann2019mvtec} and used an autoencoder-like network to learn
the representation of normal images. The network can reconstruct
anomaly-free regions of high fidelity but not for anomalous regions.  As
a result, the pixel-wise difference between the input abnormal image and
its reconstructed image reveals the region of abnormality.  A similar
idea was developed using the image inpainting technique
\cite{li2020superpixel, zavrtanik2021reconstruction}.  Traditional  
machine learning models such as support vector data description (SVDD)
\cite{tax2004support} can also be integrated with neural network, where
novel loss terms are derived to learn local image features from scratch
\cite{yi2020patch, liznerski2020explainable}.  Generally speaking,
methods without external training data either fail to provide
satisfactory performance or suffer from a slow inference speed
\cite{yi2020patch}.  This is attributed to diversified contents of
normal images. For example, the 10 object classes and the 5 texture
classes in the MVTec AD dataset are quite different. Their
capability in representing features of local regions of different images
is somehow limited.  On the other hand, over-parameterized DNN models
pretrained by external data may overfit some datasets but may not be
generalizable to other unseen contents such as new texture patterns. It
is desired to find an effective and mathematically transparent learning
method to address this challenging problem. 

\noindent
{\bf SSL and Its Applications.} SSL is an emerging machine learning
technique developed by Kuo {\em et al.} in recent years
\cite{kuo2016understanding, kuo2019interpretable, chen2020pixelhop,
chen2020pixelhop++, rouhsedaghat2021successive}. It has been applied to
quite a few applications with impressive performance.  Examples include
image classification \cite{chen2020pixelhop, chen2020pixelhop++}, image
enhancement \cite{azizi2020noise}, image compression
\cite{tseng2020interpretable}, deepfake image/video detection
\cite{chen2021defakehop}, point cloud classification, segmentation,
registration \cite{zhang2020pointhop, zhang2020pointhop++,
zhang2020unsupervised, kadam2021r}, face biometrics
\cite{rouhsedaghat2020low, rouhsedaghat2020facehop}, texture analysis
and synthesis \cite{zhang2019texture, lei2020nites}, 3D medical image
analysis \cite{liu2021voxelhop}, etc. 

\section{AnomalyHop Method}\label{sec:method}

AnomalyHop belongs to the second category of image anomaly localization
methods. Its system diagram is illustrated in Fig. \ref{fig:2_framework}.
It contains three modules: 1) feature extraction, 2) modeling
of normality feature distributions, and 3) anomaly map generation. They
will be elaborated below. 

\subsection{SSL-based Feature Extraction}\label{subsec:feature}

Deep-learning methods learn image features indirectly. Given a network
architecture, the network learns the filter parameters first by
minimizing a cost function end-to-end. Then, the network can be used to
generate filter responses, and patch features are extracted as the
filter responses at a certain layer. In contrast, the SSL framework
extracts features of image patches directly using a data-driven
approach. The basic idea is to study pixel correlations in a
neighborhood (say, a patch) and use the principal component analysis
(PCA) to define an orthogonal transform, also known as the Karhunen
Lo\`{e}ve transform (KLT).  However, a single-stage PCA transform is not
sufficient to obtain powerful features. A sequence of modifications have
been proposed in \cite{kuo2016understanding, kuo2019interpretable,
chen2020pixelhop, chen2020pixelhop++} to make the SSL framework complete.

The first modification is to build a sequence of PCA transforms in
cascade with the max pooling inserted between two consecutive stages.
The output of the previous stage serves as the input to the current
stage. The cascaded transforms are used to capture short-, mid- and
long-range correlations of pixels in an image. Since the neighborhood of
a graph is called a hop (e.g., 1-hop neighbors, 2-hop neighbors, etc.),
each transform stage is called a hop \cite{chen2020pixelhop}. However, a
straightforward cascade of multi-hop PCAs does not work properly due to
the sign confusion problem, which was first pointed out in
\cite{kuo2016understanding}. The second modification is to replace the
linear PCA with an affine transform that adds a constant-element bias vector to
the PCA response vector \cite{kuo2019interpretable}.  The bias vector is
added to ensure all input elements to the next hop are positive to avoid
sign confusion. This modified transform is called the Saab (Subspace
approximation with adjusted bias) transform. The input and the output of
the Saab transform are 3D tensors (including 2D spatial components and
1D spectral components.) By recognizing that the 1D spectral components
are uncorrelated, the third modification was proposed in
\cite{chen2020pixelhop++} to replace one 3D tensor input with multiple
2D tensor inputs. This is named as the channel-wise Saab (c/w Saab)
transform, and it greatly reduces the model size of the
standard Saab transform.
Here we employee c/w Saab transform as our feature extractor, where the output of c/w Saab transform will provide pixel-wise image local features automatically.

\subsection{Modeling of Normality Feature Distributions}\label{subsec:Gaussian} 

We propose three Gaussian models to describe the distributions of
features of normal images, which are extracted in Sec. \ref{subsec:feature}.

\subsubsection{Location-aware Gaussian Model} 

If the input images of an image class are well aligned in the spatial
domain, we expect that features at the same location are close to each
other.  We use $X_{ij}^n$ to denote the feature vector extracted from a
patch centered at location $(i, j)$ of a certain hop in the $n$th
training image. By following \cite{defard2020padim}, we model the
feature vectors of patches centered at the same location $(i,j)$ by a
multivariate Gaussian distribution, $\mathbb{N}(\mu_{ij},\Sigma_{ij})$.
Its sample mean is $\mu_{ij} = N^{-1} \sum^N_{n=1} X_{ij}^n$ and its
sample covariance matrix is
$$
\Sigma_{ij} = (N-1)^{-1} \sum^N_{n=1} (X_{ij}^n-\mu_{ij})(X_{ij}^n-\mu_{ij})^T 
+ \epsilon I,
$$
where $N$ is the number of training images of an image class, 
$\epsilon$ is a small positive number, and $I$ denotes identity matrix. The term $\epsilon I$ is added to
ensure that the sample covariance matrix is positive semi-definite. 

\subsubsection{Location-Independent Gaussian Model} 

For images of the same texture class, they have strong self-similarity.
Besides, they are often shift-invariant. These properties can be
exploited for texture-related tasks \cite{zhang2019texture,
zhang2019data,zhang2021dynamic}. For homogeneous fine-granular textures, we can use a
single Gaussian model for all local image features at each hop and call
it the location-independent Gaussian model. The model has its mean $\mu
= (NHW)^{-1} \sum_{i,j,n} X_{ij}^n$ and its covariance matrix
$$
\Sigma = (NHW-1)^{-1} \sum_{i,j,n} (X_{ij}^n- \mu_{ij})(X_{ij}^n
-\mu_{ij})^T + \epsilon I,
$$ 
where $N$ is the number of training images in one texture class, and $H$
and $W$ are pixel numbers along the height and the width of texture
images.

\subsubsection{Self-reference Gaussian Model} 

Both location-aware and location-independent Gaussian models utilize
all training images to capture the normality feature distributions.
However, images of the same class may have intra-class variations, which location-aware and location-independent Gaussian models cannot capture well. One example is the grid class in the MVTec AD dataset.
Different images may have different grid orientations and lighting
conditions. To address this problem, we train a Gaussian model with the
distribution of features from a single normal image and call it the
self-reference Gaussian. Again, we compute the sample mean as
$\mu=(HW)^{-1} \sum_{i,j} X_{ij} $ and the sample covariance matrix as
$$
\Sigma = (HW-1)^{-1} \sum_{i,j} (X_{ij}-\mu)(X_{ij} -\mu) + \epsilon I.
$$
For this setting, we only use normal images in the training set to
determine the c/w Saab transform filters. The self-reference Gaussian
model is learned from the test image at the testing time.  For more
discussion, we refer to Sec.  \ref{sec:experiments}. 

\subsection{Anomaly Map Generation and Fusion} 

With learned Gaussian models, we use the Mahalanobis distance,
$$
M(X_{ij})= \sqrt{(X_{ij}-\mu_{ij})\Sigma_{ij}^{-1}(X_{ij} -\mu_{ij})^T}, 
$$
as the anomaly score to show the anomalous level of a corresponding
patch.  Higher scores indicate a higher likelihood to be anomalous. By
calculating the scores over all locations of a hop, we form an anomaly
map at each hop for an input test image. Finally, we re-scale all anomaly
maps to the same spatial size and fuse them by weighting average to yield the final anomaly
map. 

\section{Experiments}\label{sec:experiments}

\noindent
{\bf Dataset and Evaluation Metric.} We evaluate our model on the MVTec
AD dataset \cite{bergmann2019mvtec}.  It has 5,354 images from 15
classes, including 5 texture classes and 10 object classes, collected
from real-world applications. The resolution of input images ranges from
700$\times$700 to 1024$\times$1024. The training set consists of normal
images only while the test set contains both normal and abnormal images.
The ground truth of anomaly regions is provided for the evaluation
purpose.  The area under the receiver operating characteristics curve
(AUC-ROC) \cite{dehaene2020iterative,bergmann2019mvtec} is chosen to be
the performance evaluation metric. 

\noindent
{\bf Experimental Setup and Benchmarking Methods.} First, we resize
images of different resolutions to the same resolution of
$224\times224$. Next, we apply the 5-stage Pixelhop++ to all classes for
feature extraction as shown in Fig.  \ref{fig:2_framework}. The spatial
sizes, $b \times b$, and the number, $k$, of filters at each hop are
searched in the range of $2 \le b \le 7$ and $2 \le k \le 5$,
respectively. The $2 \times 2$ max-pooling is used between hops. The
optimal hyper-parameters at each hop are class dependent. A
representative case for the leather class is given in Table
\ref{tab:hyper-parameters}. The optimal hyper-parameters of all 15
classes can be found in our
\href{https://github.com/BinWang28/AnomalyHop}{Github codes.}
We compare AnomalyHop against seven benchmarking methods. Four
of them belong to the first category that leverages external 
datasets. They are PaDiM \cite{defard2020padim}, SPADE
\cite{cohen2020sub}, DFR \cite{yang2020dfr} and CNN-FD
\cite{napoletano2018anomaly}. Three of them belong to the second category
that solely relies on images in the MVTec AD dataset. They are
AnoGAN \cite{schlegl2017unsupervised}, VAE-grad
\cite{dehaene2020iterative} and Patch-SVDD \cite{yi2020patch}. 

\begin{table}[htb]
\begin{center}
\caption{The hyper-parameters of spatial sizes and numbers of filters at
each hop for the leather class.} \label{tab:hyper-parameters}
\begin{tabular}{cccccc} \hline 
Hop Index & 1 & 2 & 3 & 4 & 5 \\ 
b & 5 & 5 & 3 & 2 & 2 \\
k & 4 & 4 & 4 & 4 & 4 \\ \hline 
\end{tabular}
\end{center}
\end{table}

\begin{table*}[htb]
\begin{center}
\caption{Performance comparison of image anomaly localization methods in
terms of AUC-ROC scores for the MVTec AD dataset, where the best results
in each category are marked in bold.} \label{exp:result}
     \resizebox{\textwidth}{!}{
    \begin{tabular}{c|c|c|c|c||c|c|c|c}
    \hline \hline
     & \multicolumn{4}{c||}{\textbf{Pretrained w/ External Data}} & \multicolumn{4}{c}{\textbf{w/o Pretraining}}\\ \hline
       & PaDiM \cite{defard2020padim} & SPADE \cite{cohen2020sub} & DFR \cite{yang2020dfr} & CNN-FD \cite{napoletano2018anomaly} & AnoGAN \cite{schlegl2017unsupervised} & VAE-grad \cite{dehaene2020iterative}  & Patch-SVDD \cite{yi2020patch} & AnomalyHop \\ \hline
      Carpet & \textbf{0.991} & 0.975 & 0.970 & 0.720 & 0.540 & 0.735 & 0.926 & \textbf{0.942}$^\ast$  \\
      Grid & 0.973 & 0.937 & \textbf{0.980} & 0.590 & 0.580 & 0.961 &  0.962 & \textbf{0.984}$^\star$ \\ 
      Leather & \textbf{0.992} & 0.976 & 0.980 & 0.870 & 0.640 & 0.925 & 0.974 & \textbf{0.991}$^\ast$ \\
      Tile & \textbf{0.941}  & 0.874 & 0.870 & 0.930 & 0.500 & 0.654 & 0.914 & \textbf{0.932}$^\ast$ \\
      Wood & \textbf{0.949} & 0.885 & 0.930 & 0.910 & 0.620 & 0.838 & \textbf{0.908} & 0.903$^\ast$  \\ \hline
      \textbf{Avg. of Texture Classes} & \textbf{0.969} & 0.929 & 0.946 & 0.804 & 0.576 & 0.823  & 0.937 & \textbf{0.950}$^{\ }$ \\ \hline
      \hline
      Bottle & 0.983 & \textbf{0.984} & 0.970 & 0.780 &  0.860 &  0.922 & \textbf{0.981} & 0.975$^\diamond$ \\
      Cable & 0.967 & \textbf{0.972} & 0.920 & 0.790 &  0.780 & 0.910 & \textbf{0.968} & 0.904$^\diamond$  \\
      Capsule & 0.985 & \textbf{0.990} & \textbf{0.990} & 0.840 & 0.840  & 0.917 & 0.958 & \textbf{0.965}$^\diamond$ \\
      Hazelnut & 0.982 & \textbf{0.991} & 0.990 & 0.720 & 0.870 & 0.976 & \textbf{0.975} & 0.971$^\diamond$  \\
      Metal Nut & 0.972 & \textbf{0.981} & 0.930 & 0.820 & 0.760 & 0.907  & \textbf{0.980} & 0.956$^\diamond$ \\
      Pill & 0.957 & 0.965 & \textbf{0.970} & 0.680 & 0.870 & 0.930 & 0.951 & \textbf{0.970}$^\diamond$  \\
      Screw & 0.985 & 0.989 & \textbf{0.990} & 0.870 & 0.800 & 0.945 & 0.957 & \textbf{0.960}$^\star$  \\
      Toothbrush & 0.988 & 0.979 & \textbf{0.990} & 0.770 & 0.900 & \textbf{0.985} & 0.981 & 0.982$^\diamond$ \\
      Transistor & \textbf{0.975} & 0.941 & 0.800 & 0.660 & 0.800 & 0.919 & 0.970 & \textbf{0.981}$^\diamond$ \\
      Zipper & \textbf{0.985} & 0.965 & 0.960 & 0.760 & 0.780 & 0.869  & 0.951 & \textbf{0.966}$^\diamond$  \\  \hline
      \textbf{Avg. of Object Classes} & \textbf{0.978} & 0.976 & 0.951  & 0.769 & 0.826 &  0.928  & \textbf{0.967} & 0.963$^{\ }$ \\ \hline \hline
      \textbf{Avg. of All Classes} & \textbf{0.975} & 0.960 & 0.949 & 0.781 & 0.743 & 0.893 & 0.957 & \textbf{0.959}$^{\ }$ \\\hline\hline
    \end{tabular}
    }
  \end{center}
\end{table*}

\noindent {\bf AUC-ROC Performance.} We compare the AUC-ROC scores of
AnomalyHop and seven benchmarking methods in Table \ref{exp:result}. As
shown in the table, AnomalyHop performs the best among all methods with
no external training data. Although Patch-SVDD has close performance,
especially for the object classes, its inference speed is significantly
slower as shown in Table \ref{exp:inference}. The best performance in
Table \ref{exp:result} is achieved by PaDiM \cite{defard2020padim} that
takes the pretrained 50-layer Wide ResNet as the feature extractor
backbone. Its superior performance largely depends on the
generalizability of the pretrained network. In practical applications,
we often encounter domain-specific images, which may not be covered by
external training data. In contrast, AnomalyHop exploits the statistical
correlations of pixels in short-, mid- and long-range neighborhoods and
obtain the c/w Saab filters based on PCA. It can tailor to a specific
application domain using a smaller number of normal images. Furthermore,
the Wide-ResNet-50-2 model has more than 60M parameters while AnomalyHop
has only 100K parameters in PixelHop++, which is used for image feature
extraction. 

\begin{figure}[htb]
\centering
\includegraphics[width=1.0\linewidth]{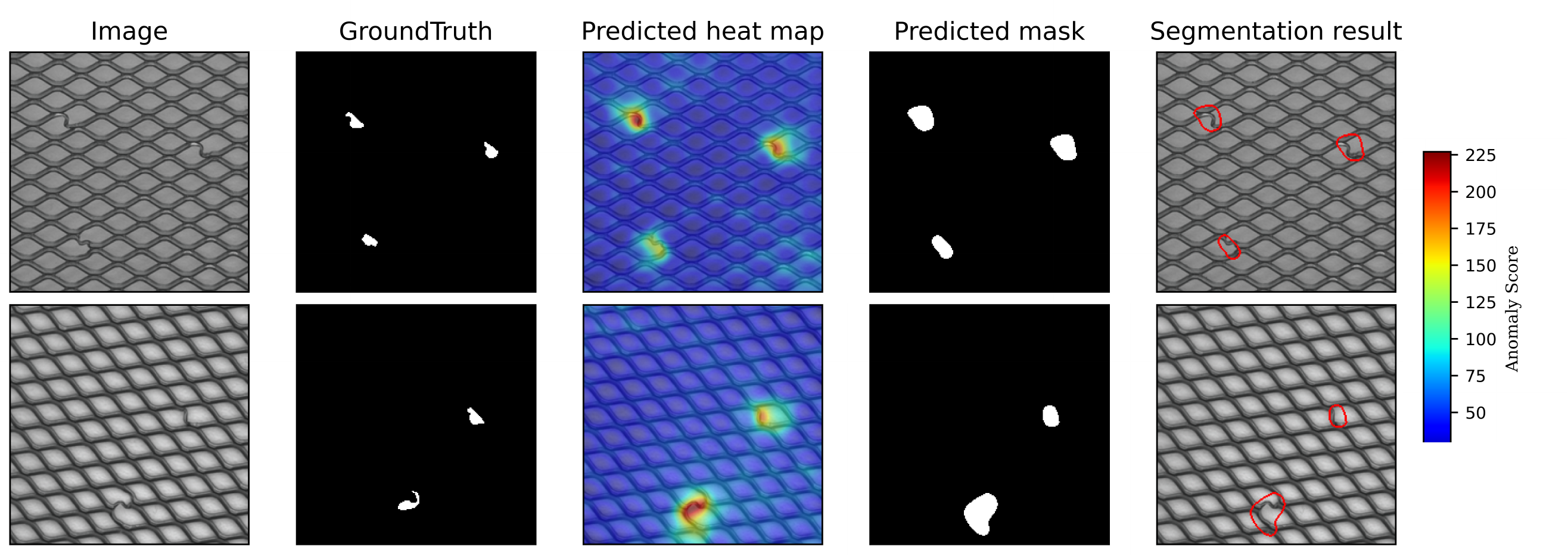}
\caption{Two anomaly grid images (from left to right): input images,
ground truth labels, predicted heatmap, predicted and segmented anomaly
regions.}\label{fig:3_grid_example}
\end{figure}

Three Gaussian models are adopted by AnomalyHop to handle different
classes in Table \ref{exp:result}.  Results obtained using
location-aware, location-independent and self-reference Gaussian models
are marked with $^\diamond$, $^\ast$, $^\star$, respectively. The object
classes are well-aligned in the dataset so that the location-aware
Gaussian model is more suitable. For texture classes (e.g. carpet and
wood classes), the location-independent Gaussian model is the most
favorable since the texture classes are usually homogeneous across the
whole image.  The location information is less relevant. The grid class
is a special one.  On one hand, the grid image is homogeneous across the
whole image.  On the other hand, different grid images have different
rotations, lighting conditions and viewing angles as shown in Fig.
\ref{fig:3_grid_example}. As a result, the self-reference Gaussian model
offers the best result. 

\noindent 
{\bf Inference Speed.} The inference speed is another important
performance metric in real-world image anomaly localization
applications. We compare the inference speed of AnomalyHop and the other
three high-performance methods in Table \ref{exp:inference}, where all
experiments are conducted with Intel I7-5930K@3.5GHz CPU. We see that
AnomalyHop has the fastest inference speed. It has a speed-up factor of
4x, 22x and 28x with respect to PaDIM, Patch-SVDD and SPADE,
respectively.  SPADE and Patch-SVDD are significantly slower because of the
expensive nearest neighbor search. For DNN-based methods, their feature
extraction can be accelerated using GPU hardware, which applies to
AnomalyHop, too.  On the other hand, image anomaly localization is often
conducted by edge computing devices in manufacturing lines. GPU could be
too expensive for this environment.  Although training complexity is
often ignored since it has to be done once, it is worthwhile to mention
that the training of AnomalyHop is very efficient. It takes only 2
minutes to train an AnomalyHop model for each class with the
above-mentioned CPU. 

\begin{table}[htb]
  \begin{center}
    \caption{Average inference time (in sec.) per image with 
    Intel i7-5930K @ 3.5 GHz CPU.}
     \label{exp:inference}
     \resizebox{0.45\textwidth}{!}{
    \begin{tabular}{|c|c|c|}
    \hline
    \textbf{Methods} & \textbf{Inference Time} & \textbf{Speed Up} \\\hline 
    SPADE \cite{cohen2020sub} & 6.80 & 1$\times$ \\ \hline
    Patch-SVDD \cite{yi2020patch} & 5.23 & 1.3$\times$ \\ \hline
    PaDiM \cite{defard2020padim} & 0.91 & 7.5$\times$ \\ \hline
    AnomalyHop & 0.24 & 28.3$\times$ \\ \hline
    \end{tabular}
    }
  \end{center}
\end{table}

\section{Conclusion and Future Work}\label{sec:conclusion}

An SSL-based anomaly image localization method, called AnomalyHop, was
proposed in this work. It is interpretable, effective and fast in both
inference and training time. Besides, it offers state-of-the-art anomaly
localization performance. AnomalyHop has a great potential to be used in
a real-world environment due to its high performance as well as low
implementation cost. 

Although SSL-based feature extraction in AnomalyHop is powerful, its
feature distribution modeling (module 2) and anomaly localization
decision (module 3) are still primitive. These two modules can be
improved furthermore. For example, it is interesting to leverage
effective one-class classification methods such as SVDD
\cite{tax2004support}, subspace SVDD \cite{sohrab2018subspace} and
multimodal subspace SVDD \cite{sohrab2021multimodal}. This is a new
topic under our current investigation. 


\bibliographystyle{IEEEtran}
\bibliography{custom}

\end{document}